\documentclass{article}

\usepackage{microtype}
\usepackage{graphicx}
\usepackage{adjustbox}
\usepackage{subfigure}
\usepackage{booktabs} 
\usepackage{amsmath}
\usepackage{amssymb}
\usepackage{multirow}
\usepackage{tablefootnote}

\usepackage{hyperref}

\usepackage[accepted]{sysml2019}

\sysmltitlerunning{Layer-compensated Pruning for Resource-constrained Convolutional Neural Networks}

\begin{document}

\twocolumn[
\sysmltitle{Layer-compensated Pruning for Resource-constrained Convolutional Neural Networks}



\sysmlsetsymbol{equal}{*}

\begin{sysmlauthorlist}
\sysmlauthor{Ting-Wu Chin}{cmu,equal}
\sysmlauthor{Cha Zhang}{ms}
\sysmlauthor{Diana Marculescu}{cmu}
\end{sysmlauthorlist}

\sysmlaffiliation{cmu}{Department of ECE, Carnegie Mellon University, Pittsburgh}
\sysmlaffiliation{ms}{Cloud and AI, Microsoft, Redmond}

\sysmlcorrespondingauthor{Ting-Wu Chin}{tingwuc@cmu.edu}

\sysmlkeywords{Meta Learning, AutoML, Model Compression, Deep Convolution Neural Networks, Pruning, Filter Pruning}

\vskip 0.3in

\begin{abstract}
    Resource-efficient convolution neural networks enable not only the intelligence on edge devices but also opportunities in system-level optimization such as scheduling. In this work, we aim to improve the performance of resource-constrained filter pruning by merging two sub-problems commonly considered, \emph{i.e.}, (i) how many filters to prune for each layer and (ii) which filters to prune given a per-layer pruning budget, into a global filter ranking problem. Our framework entails a novel algorithm, dubbed \emph{layer-compensated pruning}, where meta-learning is involved to determine better solutions. We show empirically that the proposed algorithm is superior to prior art in both effectiveness and efficiency. Specifically, we reduce the accuracy gap between the pruned and original networks from 0.9\% to 0.7\% with \textbf{8x} reduction in time needed for meta-learning, \emph{i.e.}, from 1~hour down to 7~minutes. To this end, we demonstrate the effectiveness of our algorithm using VGG, ResNet, and MobileNetV2 networks under CIFAR-10, ImageNet, and Bird-200 datasets.
\end{abstract}

]


\printAffiliationsAndNotice{\textsuperscript{*}Work started as a research intern at Microsoft.}  

\section{Introduction}
With the ubiquity of mobile and edge devices, it has become desirable to bring the performance of convolution neural networks (CNN) to the device without going through the cloud, especially due to latency and privacy considerations. However, mobile and edge devices are often characterized by stringent resource constraints, such as energy consumption and model size. Furthermore, depending on the application domain, multiply-accumulate operations (MACs) or latency may also need to be constrained. On the other hand, as the resource usage for each of the workloads (\emph{i.e.}, CNN) reduces, the system can serve more of them at once, which further creates system-level optimization opportunities. As a result, the main research question is how we design resource-constrained CNNs that are still capable to deliver good enough task performance, or prediction accuracy. This problem has been recently approached from two directions: (i) resource-constrained neural architecture search and (ii) resource-constrained model compression. On one hand, neural architecture search is a \emph{bottom-up} approach that has shown promise in finding a better solution due to the larger search space; however, it requires massive computational power to conduct the search. On the other hand, resource-constrained model compression is a \emph{top-down} approach that seeks solutions in the neighborhood of a provided pre-trained network. As a result, the final solution may be worse in terms of accuracy but it is much more efficient to find when compared to neural architecture search. We conjecture that, the latter approach provides desirable means for machine learning practitioners to take advantage from off-the-shelf, carefully-crafted deep neural networks and adapt them to user-defined resource constraints without having to search for and train a network for a long time.

Among various model compression methods, structural pruning has been widely adopted to approach the resource-constrained deep neural network optimization problem~\cite{molchanov2016pruning,he2017channel,luo2017thinet,he2018amc,gordon2018morphnet} since it provides solution that works directly on existing deep learning frameworks and hardware. Specifically, for CNNs, the structural pruning method we discuss in this paper is filter pruning~\cite{li2016pruning}.  However, resource-constrained pruning is a hard problem since (i) deciding which filter to disable such that accuracy is maximized and constraints are satisfied is a NP-hard combinatorial problem and (ii) deep neural networks are highly non-linear functions that are hard to analyze. To tackle these problems, prior art leverages approximations and considers two sub-problems: (a) how many filters to prune for each layer, which we call the \emph{layer scheduling} problem, and (b) which filters to prune, given a per-layer pruning budget, which we call the \emph{ranking} problem.

The majority of prior work focuses on the \emph{ranking} problem and proposes various heuristics to measure the importance of each filter~\cite{jiang2018efficient,mao2017exploring,he2017channel,luo2017thinet,yoon2018filter,li2016pruning} while addressing \emph{layer scheduling} by either employing manually designed rules based on experience or simply pruning the network uniformly by a fixed percentage across layers. To achieve effective and efficient filter pruning, there is a need for algorithms that determine the layer schedule without humans in the loop. Pioneering this direction,~\cite{he2018amc} approach this problem with reinforcement learning to learn an agent for deciding how many filters to prune for each layer given a resource constraint. While achieving better results than a hand-tuned policy, we note that such a formulation is too time-consuming if the goal is the Pareto frontier of the design space of interest. We argue that traversing the Pareto frontier efficiently is a critical feature desired by machine learning practitioners since constraints are not always known \emph{a priori} when building a system running convolution neural network models among other workloads.

In this work, we first theoretically derive the pruning problem such that the \emph{layer scheduling} problem and the \emph{ranking} problem are treated as a unified problem with a na\"ive solution. Then, based on the analysis of the derivation, we introduce a novel formulation that takes the approximation errors into account to further improve the na\"ive solution. Specifically, we leverage meta-learning to learn a set of latent variables that compensate for the approximation error, and we call it the \emph{layer-compensated pruning} algorithm. Overall, our contributions are as follows:
\begin{itemize}
    \item We define the pruning problem from a theoretical standpoint and connect to prior solutions throughout the derivations.
    \item We propose a novel, effective, and efficient algorithm, dubbed \emph{layer-compensated pruning}, which improves prior art by learning to compensate for the approximation error incurred in the derivation in a layer-wise fashion. Specifically, we achieve slight better results 8x faster. 
    \item In our general formulation, we show that layer-compensated pruning can improve various commonly-adopted heuristic metrics such as $\ell_1$, $\ell_2$ of weights, and $1^{st}$-order Taylor approximation.
    \item We conduct comprehensive analysis to justify our findings on already-small deep neural networks, \emph{i.e.}, ResNet and MobileNetV2, using three datasets, \emph{i.e.}, CIFAR-10, ImageNet, and Bird-200.
\end{itemize}


\section{Related Work}
For resource-constrained DNN design, there are generally two approaches: bottom-up and top-down.
\subsection{Bottom-up Resource-constrained DNN Design}
Bottom-up approaches try to build the neural network from ground-up while incorporating some awareness of the resources. These types of approaches are often called multi-objective or platform-aware neural architecture search~\cite{hsu2018monas,zhou2018resource,tan2018mnasnet,dong2018dpp}. Within this domain, every approach has a different search space, learning algorithm, and decision space. For example, \cite{zhou2018resource} used a policy gradient algorithm to learn two controllers: one is in charge of scaling the number of neurons in an existing layer and the other removes or inserts new operations on top of the current layer given the network embedding. Similarly, \cite{hsu2018monas} used policy gradient to learn a controller that suggests the hyper-parameters for the target network given the network embedding, but with a smaller search space that tunes the hyper-parameters of an existing architecture (\emph{e.g.}, number of filters in convolution and growth rate in CondenseNet~\cite{huang2017condensenet}). On the other hand, \cite{tan2018mnasnet} relied on a search space that covers connections and operations within a cell and determines how many stacked cells should form a block. Lastly, \cite{dong2018dpp} proposed to learn a surrogate function that predicts the accuracy of network candidates, which makes the traversal of the Pareto frontier efficient. Although bottom-up approaches enable a promising solution with larger search space, they are much more time-consuming compared to top-down approaches.
\subsection{Top-down Resource-constrained DNN Design}
We consider pruning and quantization top-down approaches since they start with an existing neural network and try to trim down connections or precision. While there is emerging work toward network quantization~\cite{khoram2018adaptive} in a resource-constrained setting, structural pruning is often adopted as the solution for the problem~\cite{gordon2018morphnet,luo2017thinet,he2017channel,molchanov2016pruning,he2018amc,yang2018end,jiang2018efficient,he2018soft,hu2016network,liu2017learning,li2016pruning,lin2018accelerating,wen2016learning} given its fine-grained control capability over the constraints. We group solutions to structural pruning into the following three categories. 
\paragraph{Joint Optimization}
In this line of work, current approaches try to jointly optimize for model weights $\mathbf{\Theta}$ as well as the filter mask $\mathbf{z}$. A common approach is to let the scaling factor of the batch normalization layer that follows the convolution layer act as the mask $\mathbf{z}$ and add $\lambda$-weighted regularization terms to suppress the scaling factor of the batch normalization layer. As a result, standard training procedures using stochastic gradient descent optimize both $\mathbf{\Theta}$ and $\mathbf{z}$ jointly~\cite{ye2018rethinking,liu2017learning,gordon2018morphnet}. On the other hand,~\cite{louizos2017bayesian} formulate the loss function from a Bayesian standpoint while~\cite{dai2018compressing} derive the loss function from an information-theoretic standpoint. However, we note that such approaches count on the non-intuitive tuning knobs to traverse the Pareto optimal.
\paragraph{Local Ranking with Layer Scheduling}
In this line of research, resource-constrained filter pruning is done by solving two sub-problems: (a) decide how many filters to prune for each layer such that the constraint is satisfied, which we call \emph{layer scheduling}, and (b) decide which filters to prune, given a layer schedule. While most of the prior art~\cite{jiang2018efficient,mao2017exploring,he2017channel,luo2017thinet,yoon2018filter,li2016pruning} decides the layer schedule manually based on experience, we argue that the layer schedule directly affects the resource usage of the resulting network. As a result, it is not scalable to rely on experts to determine the layer schedule so as to traverse the Pareto frontier. Although \cite{he2018amc} provide a good and scalable (compared to human-in-the-loop) solution for resource-constrained pruning, the reinforcement learning agent generates the layer schedule sequentially for each layer, which is inefficient and presumably harder to learn for deeper networks due to the credit assignment problem~\cite{sutton1998reinforcement}. 
\paragraph{Global Ranking with Single-filter Pruning}
In this category, the filters of the entire network are ranked together according to some heuristic metrics and pruning is an iterative process that has three steps: rank filters globally, greedily prune one filter at a time, and fine-tune the network. The process continues until resource constraints are satisfied, which makes the traversal along different values of constraints intuitive. \cite{molchanov2016pruning} propose to leverage first-order Taylor approximation for the ranking and progressively prune one filter at a time, while \cite{theis2018faster} later propose to leverage Fisher information for ranking. However, pruning one filter at a time is not scalable when the number of filters is large, as in the case of MobileNetV2~\cite{sandler2018mobilenetv2} which has 17k filters. While our work is also considered a global ranking approach, we conduct multi-filter pruning instead of single-filter pruning, and we find multi-filter pruning to be more effective and more efficient.

\section{Problem Formulation}
We are interested in solving the problem of resource-constrained filter pruning. Formally, the optimization problem to consider is:
\begin{align}
    \begin{split}
        \min_{\mathbf{z},\mathbf{\Theta}}~\left|\mathbb{L}(\mathbf{\Theta}^*)-\mathbb{L}(\mathbf{\Theta}\odot \mathbf{z})\right|\\
        \text{s.t.}~C(\mathbf{z}) \le \zeta.
    \end{split}\label{eq:hard_problem}
\end{align}
In the objective function, $\mathbb{L}(\mathbf{\Theta}) = \mathbb{E}_{(x,y)\sim D}~L(f(x|\mathbf{\Theta}),y)$, where $x$ and $y$ are the input and label, respectively, $D$ is the data distribution we care about, $L(\cdot)$ is the loss function, $f$ is the DNN model, $\mathbf{\Theta}=\{\mathbf{\theta}_1, \cdots, \mathbf{\theta}_K\}$ and $\mathbf{\Theta}^*=\{\mathbf{\theta}_1^*, \cdots, \mathbf{\theta}_K^*\}$ are the target and pre-trained model weights, which are $K-$dimensional array of vectors with each dimension representing the weights of a filter. $\mathbf{z}=\{z_1, \cdots, z_K\}$ is a $K-$dimensional indicator vector representing if the corresponding filter $k$ is pruned ($z_k=0$) or not ($z_k=1)$. 

On the constraint side, $C(\cdot)$ is the target resource usage evaluation function and $\zeta$ is the desired resource constraint. For ease of notation, we denote by $\left|\mathbb{L}(\mathbf{\Theta}^*)-\mathbb{L}(\mathbf{\Theta}\odot \mathbf{z})\right|$ the loss difference caused by $\mathbf{z}$. We approach this optimization problem with the commonly adopted optimization framework~\cite{he2018amc,he2017channel,luo2017thinet,li2016pruning,yang2018end,yoon2018filter,mao2017exploring,he2018soft,jiang2018efficient,lin2018accelerating} as shown in Algorithm~\ref{alg:rcp}.
\begin{algorithm}[t]
   \caption{Resource Constrained Pruning Framework}
   \label{alg:rcp}
\begin{algorithmic}[1]
   \STATE {\bfseries Input:} pre-trained model $\mathbf{\Theta}^*$, constraint $\zeta$, iteration count $t$, resource usage function $C(\cdot)$
   \STATE {\bfseries Output:} pruned model $\mathbf{\Theta}^{(t)}$
   \STATE $\mathbf{\Theta}^{(0)} = \mathbf{\Theta}^*$
   \STATE $\tilde{\zeta}^{(0)} = C(\mathbf{1})$
   \FOR{$i=1$ {\bfseries to} $t$}
   \STATE {\bfseries Pruning:} $\mathbf{z}^{(i)}$ = solve (\ref{eq:hard_problem}) with fixed $\mathbf{\Theta}=\mathbf{\Theta}^{(i-1)}$ and constraint $\tilde{\zeta}^{(i-1)}$
   \STATE {\bfseries Tuning:} $\mathbf{\Theta}^{(i)}$ = solve (\ref{eq:hard_problem}) with fixed $\mathbf{z}=\mathbf{z}^{(i)}$
   \STATE {\bfseries Constraint tightening:} $\tilde{\zeta}^{(i)} \in [\zeta, \tilde{\zeta}^{(i-1)})$
   \ENDFOR
\end{algorithmic}
\end{algorithm}

Note that solving $\mathbf{z}$ optimally in each iteration of Algorithm~\ref{alg:rcp} (line 5) is a combinatorial problem that takes O($2^K$) evaluations of the objective. In practice, greedy approximation is often adopted to find a sub-optimal $\mathbf{z}$. 

\subsection{Greedy Single-filter Pruning}
To solve Algorithm~\ref{alg:rcp} (line 5) in a greedy fashion, prior work assumes that $\mathbf{z}^{(i)}$ will prune one additional filter on top of $\mathbf{z}^{(i-1)}$~\cite{molchanov2016pruning,theis2018faster}. That is, they seek to optimize: 
\begin{align}
    \begin{split}
        \min_{j\in \mathbb{R}(\mathbf{z}^{(i-1)})}~\left|\mathbb{L}(\mathbf{\Theta}^{(i-1)})-\mathbb{L}(\mathbf{\Theta}^{(i-1)}\odot \mathbf{z}_{-j})\right|\\
        \text{s.t.}~C(\mathbf{z}^{(i)}) \le \zeta, 
    \end{split}\label{eq:greedy_opt}
\end{align}
where $\mathbf{z}_{-j}$ represents the pruning of the $j^{th}$ filter, i.e., $z_j=0$ and $z_k=1, k=1,\cdots,K, k\neq j$. $\mathbb{R}(\mathbf{z}^{(i-1)})$ is the set of remaining non-zero elements in $\mathbf{z}^{(i-1)}$. And 
\begin{align}
    \mathbf{z}^{(i)} = \mathbf{z}^{(i-1)}\odot \mathbf{z}_{-j}
\label{eq:greed_z}
\end{align}
In the remainder of this section, we will stop using superscripts whenever there is no confusion. 

Computing the loss difference term in equation~(\ref{eq:greedy_opt}) is nontrivial as a data set needs to be prepared to compute the change in loss function and one needs to evaluate the difference $|\mathbb{R}(\mathbf{z}^{(i-1)})|$ times to proceed in each iteration. A common practice is to define a metric to quantify the loss difference. That is: 
\begin{align}
    \begin{split}
        \left|\mathbb{L}(\mathbf{\Theta})-\mathbb{L}\left(\mathbf{\Theta}\odot\mathbf{z}_{-j})\right)\right| = M_j(\mathbf{\Theta}) + \epsilon_{M_j}
    \end{split}\label{eq:metric_approx},
\end{align}
where $M_j(\mathbf{\Theta})$ can be easy to compute based on the magnitude of the model parameters $\mathbf{\Theta}$ and $\epsilon_{M_j}$ is the approximation error incurred with metric approximation.

In the literature, various metric approximations $M_j(\mathbf{\Theta})$ have been proposed. For example, past work has used $\ell_2$ of filter weights~\cite{he2018soft,yang2018netadapt}, $\ell_1$ of filter weights~\cite{mao2017exploring,li2016pruning}, $\ell_2$ of filter weights and filter weights of the next layer~\cite{jiang2018efficient}, first-order Taylor expansion on the loss~\cite{molchanov2016pruning,lin2018accelerating}, Fisher information of the loss~\cite{theis2018faster}, and the variance of the max activation~\cite{yoon2018filter}.

\subsection{Greedy Multi-filter Pruning}
Greedy single-filter pruning takes  $JK-\frac{J(J-1)}{2}$ evaluations on the objective and fine-tuning to prune $J$ filters from a pretrained model with $K$ filters. To prune more filters at a time, or even perform one-shot pruning, it is common to approximate the loss difference caused by a set of pruned filters with addition~\cite{luo2017thinet,li2016pruning,yang2018end,yoon2018filter,he2018soft,he2018amc,he2017channel,mao2017exploring,jiang2018efficient,lin2018accelerating,yang2018netadapt}. Formally,
\begin{align}
    \begin{split}
        \left|\mathbb{L}(\mathbf{\Theta})-\mathbb{L}\left(\mathbf{\Theta}\odot \mathbf{z}_{-\{j_1,\cdots,j_p\}}\right)\right| =\\
        \sum_{j\in\{j_1,\cdots,j_p\}}\left|\mathbb{L}(\mathbf{\Theta})-\mathbb{L}\left(\mathbf{\Theta}\odot \mathbf{z}_j\right)\right| + \varepsilon =\\
        \sum_{j\in\{j_1,\cdots,j_p\}} (M_j(\mathbf{\Theta}) + \epsilon_{M_j}) + \varepsilon,
    \end{split}\label{eq:additive_approx}
\end{align}
where $j_1,\cdots,j_p$ are $p$ newly pruned filters, $\varepsilon$ is the approximation error for addition. 

If the approximation errors $\epsilon_{M_j}$ and $\varepsilon$ are both negligible, solving $\mathbf{z}$ in line~5 of Algorithm~\ref{alg:rcp} is then equivalent to solving the following problem:
\begin{align}
    \begin{split}
        \min_{\mathbf{z}}~\sum_{j=1}^K M_j(\mathbf{\Theta}) z_j\\
        \text{s.t.}~C(\mathbf{z}) \le \zeta.
    \end{split}\label{eq:notreally_knapsack}
\end{align}
\begin{figure*}[h]
    \centering
    \includegraphics[width=\linewidth]{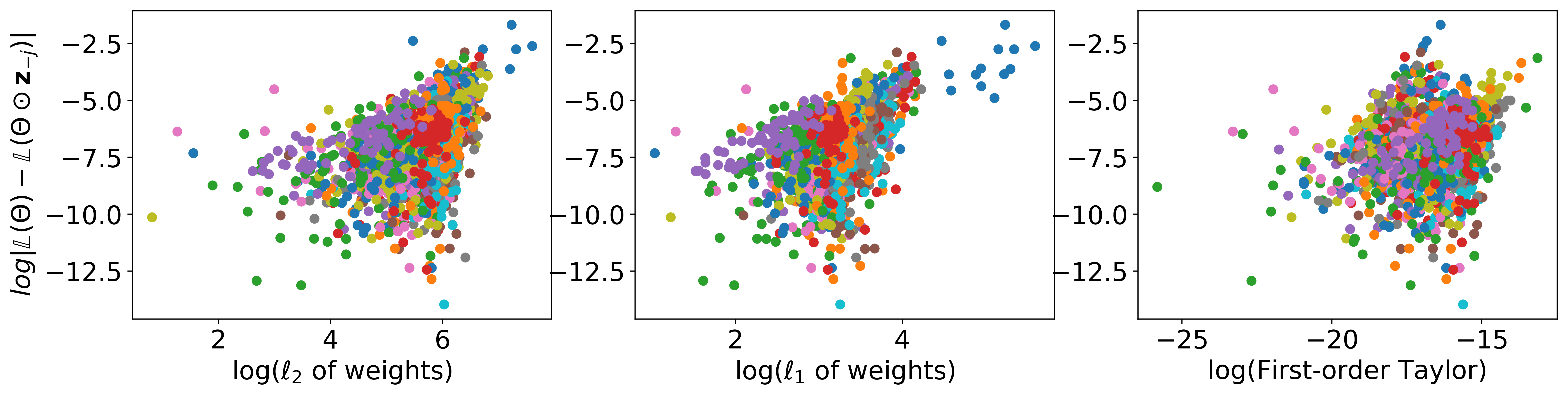}
    \caption{Calculated loss difference versus various heuristic metrics in log-scale. Each point is a filter of ResNet-56 trained on CIFAR-10. Different colors represent filters on different layers.}
    \label{fig:merror}
\end{figure*}

\paragraph{Local Ranking with Layer Scheduling}
Prior art~\cite{luo2017thinet,li2016pruning,yoon2018filter,he2018soft,he2018amc,he2017channel,jiang2018efficient,lin2018accelerating,yang2018netadapt} adopts the aforementioned assumption and approaches equation (\ref{eq:notreally_knapsack}) by introducing the layer scheduling variable into the picture:
\begin{align}
    \begin{split}
        \min_{\mathbf{z}}~\sum_{j=1}^K M_j(\mathbf{\Theta}) z_j\\
        \text{s.t.}~||G(\mathbf{z})_l||_0 = LS_l,~~l=1,\dots,L\\
        C(\mathbf{z}) \le \zeta,
    \end{split}\label{eq:layer_schedule}
\end{align}
and solving it greedily with layer-wise top-\emph{k} selection. In equation (\ref{eq:layer_schedule}), $G(\cdot)_l$ retrieves the filters in the $l^{th}$ layer and $LS_l$ is the number of filters to be kept for the $l^{th}$ layer. We note that, in this fashion, the layer schedule directly detemines whether the resource constraint is met or not.

\paragraph{Na\"ive Pruning}
Equation (\ref{eq:notreally_knapsack}) can certainly also be solved with a global greedy pruning algorithm without layer scheduling. That is, greedily prunes the filter that has least loss difference $M_j(\mathbf{\Theta})$ until the constraint is satisfied, which we call the \emph{na\"ive pruning}.

\section{Layer-Compensated Pruning}\label{sec:lcd}
One might wonder in the above discussion how acceptable is the assumption that the errors $\epsilon_{M_j}$ and $\varepsilon$ are both negligible. We conducted an experiment as shown in Fig.~\ref{fig:merror} where we plot the heuristic metrics on the \emph{x-}axis and the calculated loss difference on the \emph{y-}axis for each of the filter for three common metrics with different colors represent different layers. It is obvious that $\epsilon_{M_j}$ is not negligible in practice, specifically, for those filters that have $e^{-2}$ in $\ell_2$ of weights, the loss difference ranges from $e^{-12.5}$ to $e^{-2.5}$. Since $\epsilon_{M_j}$ is not negligible, equation (\ref{eq:notreally_knapsack}) should include the error term as follows:
\begin{align}
    \begin{split}
        \min_{\mathbf{z}}~\sum_{j=1}^K \left(M_j(\mathbf{\Theta})+\epsilon_{M_j}\right) z_j\\
        \text{s.t.}~C(\mathbf{z}) \le \zeta.
    \end{split}\label{eq:error_opt}
\end{align}
To avoid computing $\epsilon_{M_j}$, we conjecture that good approximation error estimations lead to better solutions from na\"ive pruning. Furthermore, we leverage a heuristic that treats the approximation errors $\epsilon_{M_j}$ to be identical for filters in the same layer. As a result, we propose a set of layer-dependent latent variables $\beta_l, l=1,\cdots,L$ that represent the error estimation to be solved. Here $L$ is the number of layers in the network. Namely, before solving equation (\ref{eq:error_opt}), we first solve the following equation to obtain the per-layer error estimation:
\begin{align}
    \begin{split}
       \min_{\beta}~\left|\mathbb{L}(\mathbf{\Theta})-\mathbb{L}\left(\mathbf{\Theta}\odot \mathbf{z}_{-\mathbb{P}}\right)\right|\\
        \mathbb{P} = Na\text{\emph{\"i}}ve\left(\left[M_j(\mathbf{\Theta})+\beta_{l(j)}\right]_{\forall j}, \zeta \right),
    \end{split}\label{eq:latent}
\end{align}
where $l(j)$ retrieves the layer index that filter $j$ belongs to and $\mathbb{P}$ is a set of filter indexes to be pruned, and $Na$\emph{\"i}$ve$ represents the na\"ive pruning. Intuitively speaking, we need to find error compensations $\beta$ such that the solution from na\"ive pruning has the minimum loss difference. Once $\beta$ is obtained, we then replace $\epsilon_{M_j}$ in equation~(\ref{eq:error_opt}) with $\beta$ and solve it with na\"ive pruning. While this is still an approximation, we show in later experiments that this approach produces networks with smaller loss difference compared to solving equation (\ref{eq:notreally_knapsack}) na\"ively on the unseen testing dataset.

\subsection{Learning the Layer-wise Compensations}\label{sec:lc}
Equation (\ref{eq:latent}) can be approached by derivative-free optimization meta-heuristic algorithms such as genetic algorithms, evolutionary strategies, and evolutionary algorithms; we leverage the regularized evolutionary algorithm proposed in~\cite{real2018regularized} for its effectiveness in the neural architecture search problem.
In our regularized evolutionary algorithm, we first generate a pool of  candidates ($\beta$s), then repeat the following steps: (i) sample a subset from the candidates, (ii) identify the fittest candidate, (iii) generate a new candidate by mutating the fittest candidate, and (iv) replace the oldest candidate in the pool with the generated one. We define the fittest candidate as the one with the minimum objective value in equation (\ref{eq:latent}).
We initialize the pool by sampling $\beta$ from $L$ normal distributions, \emph{i.e.},~$\beta_l~\sim~N(0,\sigma_{ml})~~\forall~l$. We denote $\sigma_{ml}$ as the standard deviation of the metric of filters at layer $l$. For mutation, we randomly select subset $S$ of current $\beta$ to be perturbed by noise $h \sim N(0,\alpha \sigma_{ml})$, \emph{i.e.}, $\beta_l=\beta_l+h,~~\forall~l\in S$, where $d=|S|$ is the hyper-parameter that controls the exploration step of the search and $\alpha$ is the step size which can be gradually decreased to reduce exploration in the later stage of the optimization. 



\section{Evaluations}
\subsection{Datasets and Training Setting}
Our work is evaluated on various benchmark including CIFAR-10~\cite{krizhevsky2009learning}, ImageNet~\cite{russakovsky2015imagenet}, and Birds-200~\cite{wah2011caltech}. The first one is a standard image classification dataset that consists of 50k training images and 10k testing images with total 10 classes to be classified while the second dataset is a large scale image classification dataset that includes 1.2 million training images and 50k testing images with 1000 classes to be classified. On the other hand, we benchmark the layer-compensated pruning on a transfer learning setting as well since in practice, we want a small and fast model on some target datasets rather than ImageNet, and hence, we use the Birds-200 dataset that consists of 6k training images and 5.7k testing images covering 200 species of bird.

For CIFAR-10, the training parameters for the baseline models follow prior work~\cite{he2018soft}, which uses stochastic gradient descent with nesterov, 0.1 initial learning rate and drop by 10x at epochs 60, 120, and 160, and train for 200 epochs in total. For pruning, we keep all training hyper-parameters the same but change the initial learning rate to 0.01 and only train for 60 epochs. We drop the learning rate by 10x at corresponding epochs, \emph{i.e.}, epochs 18, 36, and 48. On the other hand, we use a pre-trained model on ImageNet and when fine-tuning, we use an initial learning rate of $1e^{-4}$ which is dropped by 10x at epoch 20. We only train for 30 epochs similar to prior work~\cite{he2018amc}. For all transfer learning datasets, we fine-tune the models that are pre-trained on ImageNet on the target datasets with 0.001 learning rate for the last layer and 0.0001 for other layers; we train for 60 epochs and drop the learning rate by 10x at epoch 48 to obtain the baseline model on the target dataset.

\subsection{Implementation Details}\label{sec:impl}
\paragraph{MobileNetV2 for CIFAR-10}
We made some changes to the original MobileNetV2 architecture design~\cite{sandler2018mobilenetv2} to adapt it from ImageNet with 224x224 input image size to CIFAR-10 with 32x32 input image size. Specifically, we change the convolution stride from two to one for block two and block four. With the changes, the final layer produces 8x8 feature maps, which is also the case for the ResNet that is designed for CIFAR-10~\cite{he2016deep}.
\paragraph{Heuristic Metrics}
We mainly consider $\ell_2$ of weights unless noted otherwise and we conduct an ablation study in Section~\ref{sec:eval} for other ranking metrics as well. Specifically, $\ell_1$ of weights of a filter $F_i$ is calculated by $M_i = ||F_i||_1$, which is the absolute summation across all three dimensions of a filter~\cite{li2016pruning}, \emph{i.e.}, input channels, kernel width, and kernel height. Similarly, $\ell_2$ is $M_i = ||F_i||_2$, which is the squared summation over all three dimensions~\cite{he2018soft}. Lastly, first-order Taylor approximation is calculated as $M_i = |\text{avg}(\nabla_{F_i} L \odot F_i)|$~\cite{molchanov2016pruning}.
\paragraph{Pruning Residual Connections}
We note that pruning residual connections is tricky since pruning complementary filters for operands of the residual addition will result in an output with the same dimension. Concretely, assuming we want to prune two six-filters kernels that are added together by a residual connection, if we prune the first three filters of one operand and the last three filters of the other operand, the output of the addition would still be six channels. To avoid this complication, we follow prior work~\cite{gordon2018morphnet} and group filters that are added by residual together. That is, either prune or do not prune together. We use addition to calculate the metric for a group.
\paragraph{Limiting Pruning Budget}
We limit the number of filters left for each layer to be at least 10\% of the original number to avoid extreme pruning and we experiment with various values later in Section~\ref{sec:limit}. We note that this is also adopted by~\cite{he2018amc} where they use 20\% as the limiting budget for each layer.
\paragraph{Evolutionary Algorithm}
For the evolutionary algorithm that searches for $\beta$, we set the total number of iterations to 336 and the size of candidate pool to 64 so that the total number of candidate seen is 400, which is the same as in prior work~\cite{he2018amc}. We arbitrarily select $d$ to be $0.1 L$, which means randomly selects a tenth of the $\beta$ to perturb for each mutation. We linearly reduce the step size $\alpha$ to avoid moving back-and-forth around the local optima, \emph{i.e.}, $\alpha = \frac{\text{cur\_iter}}{\text{total\_iter}}$. We randomly sample 3k images from the \emph{training set}, which follows~\cite{he2018amc}, to evaluate the objective in (\ref{eq:latent}) for each candidate.
\subsection{Analysis on CIFAR-10}\label{sec:eval}
For our methods on CIFAR-10, we conduct one-shot pruning. That is, we set the constraint to be the target constraint in the first iteration of Algorithm~\ref{alg:rcp}.
\begin{figure*}[h]
    \centering
    \includegraphics[width=0.9\linewidth]{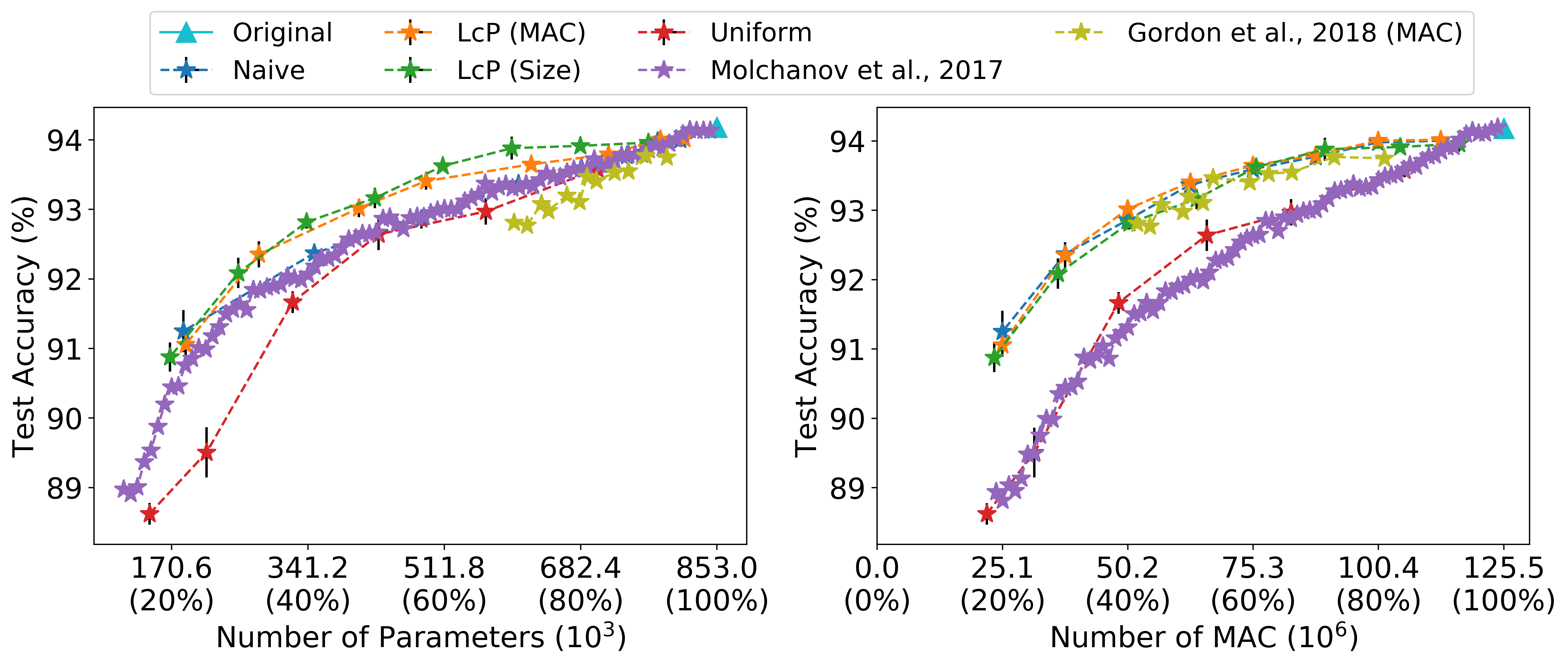}
    \caption{The Pareto frontier of pruning ResNet-56 on CIFAR-10 using various of methods. Uniform means uniformly pruning a fixed percentage out of every layer. With methods that have error bars (na\"ive, uniform, and layer-compensated), we average across three trials and plot the mean and standard deviation.}
    \label{fig:pareto}
\end{figure*}
\begin{table*}[h]
\caption{Comparison with prior art. We group methods into sections according to different MAC operations. Values for our approaches are averaged across three trials and we report the mean and standard deviation. We use bold face to denote the best numbers.}
\vskip 0.15in
\begin{center}
\begin{small}
\begin{sc}
\begin{adjustbox}{max width=1\textwidth}
\begin{tabular}{c|cccc}
\toprule
Network & Method & Acc. (\%) & MAC ($10^6$) & Cost of Pareto frontier\\
\midrule
\multirow{12}{*}{ResNet-56}&\cite{li2016pruning} & 93.04 $\xrightarrow{}$ 93.06 & 90.9 (72.4\%) & $(T_{human} + T_{network})\times N$\\
&\cite{molchanov2016pruning}$^*$ & \textbf{94.18} $\xrightarrow{}$ 93.21 & 90.8 (72.4\%)&$T_{shortFT}\times J$\\
&Na\"ive & \textbf{94.18} $\xrightarrow{}$ 93.77$\pm$0.06 & \textbf{87.8} (70\%)&$T_{network}\times N$\\
&LcP (Ours) & \textbf{94.18} $\xrightarrow{}$ \textbf{93.79$\pm$0.11} & \textbf{87.8} (70\%)&$(T_{meta}+T_{network})\times N$\\
\cline{2-5}
&\cite{gordon2018morphnet}$^*$ & \textbf{94.18} $\xrightarrow{}$ 93.51 & 77.8 (62\%)&$(T_{\lambda}+T_{network})\times N$\\
&Na\"ive & \textbf{94.18} $\xrightarrow{}$ 93.59$\pm$0.04 & \textbf{75.2} (60\%)&$T_{network}\times N$\\
&LcP (Ours) & \textbf{94.18} $\xrightarrow{}$ \textbf{93.65$\pm$0.06} & 75.3 (60\%)&$(T_{meta}+T_{network})\times N$\\
\cline{2-5}
&\cite{he2017channel} & 92.8 $\xrightarrow{}$ 91.8 & 62.7 (50\%)& $(T_{human} + T_{network})\times N$\\
&\cite{he2018amc} & 92.8 $\xrightarrow{}$ 91.9 & 62.7 (50\%)& $(T_{meta} + T_{network})\times N$ \\
&\cite{he2018soft} & 93.59$\pm$0.58 $\xrightarrow{}$ 93.35$\pm$0.31 & \textbf{59.4} (47.4\%)& $T_{network}\times N$ \\
&Na\"ive & \textbf{94.18} $\xrightarrow{}$ 93.37$\pm$0.05 &   62.6 (50\%)&$T_{network}\times N$\\
&LcP (Ours) & \textbf{94.18} $\xrightarrow{}$ \textbf{93.41$\pm$0.12} & 62.7 (50\%)&$(T_{meta}+T_{network})\times N$\\
\midrule
\multirow{5}{*}{VGG-13}&\cite{louizos2017bayesian} & 91.9 $\xrightarrow{}$ 91.4 & 141.5 (45.1\%) & $(T_{\lambda}+T_{network})\times N$\\
&\cite{louizos2017bayesian} & 91.9 $\xrightarrow{}$ 91 & 121.9 (38.9\%) & $(T_{\lambda}+T_{network})\times N$\\
&\cite{dai2018compressing} & 91.9 $\xrightarrow{}$ 91.5 & 70.6 (22.5\%) & $(T_{\lambda}+T_{network})\times N$\\
&Na\"ive & 91.9 $\xrightarrow{}$ 91.78$\pm$0.27 & \textbf{70.1} (22.4\%) & $T_{network}\times N$\\
&LcP (Ours) & 91.9 $\xrightarrow{}$ \textbf{92.38}$\pm$0.19 & 70.3 (22.4\%) &$(T_{meta}+T_{network})\times N$\\
\bottomrule
\end{tabular}\label{table:cifar10}
\end{adjustbox}
\end{sc}
\end{small}
\end{center}
~$^*$Our implementation
\vskip -0.1in
\end{table*}
\paragraph{Effectiveness}
As shown in Fig.~\ref{fig:pareto}, we find that when targeting the MAC constraint, the \emph{na\"ive pruning} outperforms the uniform pruning baseline and \emph{single-filter pruning}~\cite{molchanov2016pruning} by a large margin and is slightly better than a \emph{joint optimization} technique that is optimized for MAC constraint~\cite{gordon2018morphnet}. Theoretically speaking, single-filter pruning should have smaller approximation error and leads to better solutions, however, we find that due to the small learning rate for tuning between the iterations, it is easy for the network to get stuck at a local optimal. We postulate that each point obtained by single-filter pruning could be further improved by extra fine-tuning with a larger learning rate. Beyond the na\"ive approach, the layer-compensated pruning (LcP) is able to obtain a better Pareto frontier than all other methods and it is effective under both model size and number of MAC operations constraints. We also compare with prior art and summarize them in Table~\ref{table:cifar10}, where LcP performs favorably compared to prior art. We note that among the prior methods that we are comparing, five of them~\cite{molchanov2016pruning,gordon2018morphnet,he2018amc,louizos2017bayesian,dai2018compressing} target the layer scheduling problem directly or indirectly while others focus on the complementary problems such as the ranking problem~\cite{he2017channel,li2016pruning} and the optimization process~\cite{he2018soft}, \emph{i.e.}, line 6 of Algorithm~\ref{alg:rcp}. In particular, compared to prior work~\cite{he2018amc} that uses reinforcement learning to learn the layer schedule for each layer, our algorithm produces lower accuracy degradation with a stronger baseline model.

\paragraph{Efficiency}
We include the Pareto frontier traversal time cost in Table~\ref{table:cifar10} where $T_{human}$ is the time needed for human experts to design the layer schedule, $T_{\lambda}$ is the trade-off parameter tuning time for joint optimization approaches, $T_{shortFT}$ is the time it takes for short term fine-tuning for single-filter pruning, $T_{meta}$ is the time it takes for meta-learning to learn the layer schedule given a target constraint value, $T_{network}$ represents the time to train a pruned network, $N$ represents the number of points of interest on the Pareto frontier, and $J$ is the number of filters needed to be pruned to achieve the lowest constraint value of interest in single-layer pruning method. First, single-filter pruning~\cite{molchanov2016pruning} requires filters to be pruned one by one and conducts fine-tuning between iterations, which incurs huge overhead to obtain networks with stringent resource usage and to prune networks with large filter numbers. Second, the reinforcement learning approach proposed in prior work~\cite{he2018amc} requires the reinforcement learning agent to be learned for each constraint values considered, where each learning takes 1 hour on GeForce GTX TITAN Xp GPU for CIFAR-10. Last, although the joint optimization techniques~\cite{gordon2018morphnet,louizos2017bayesian,dai2018compressing} are efficient for Pareto frontier traversal since there is no separate optimization for $\mathbf{\Theta}$ and~$\mathbf{z}$, controlling the trade-off of resource usage and accuracy in joint optimization methods is not intuitive and requires several trial-and-error to achieve the target constraint. Specifically, it is necessary to tune the $\lambda$ regularizer to traverse the Pareto frontier, \emph{i.e.}, we tune the $\lambda$ from $2\times 10^{-9}$ all the way to $13\times 10^{-9}$ to obtain points for prior work~\cite{gordon2018morphnet} in Fig.~\ref{fig:pareto}. Moreover, such a trade-off knob is affected by the learning rate, that is, different learning rates result in networks with different resource usage while keeping the trade-off knob fixed. In comparison, LcP has an intuitive traversal knob, \emph{i.e.}, MAC operations or model size. Therefore, it is more efficient for traversing the Pareto frontier compared to joint optimization approaches since it eliminates the need for tuning learning rate and the trade-off hyper-parameters. Compared to the reinforcement learning-based approach~\cite{he2018amc}, our algorithm only requires \textbf{7~minutes} on a single GeForce GTX 1080 Ti GPU, which is comparable or slightly worse than GeForce GTX TITAN Xp GPU, to solve equation~(\ref{eq:latent}) while observing the same number of candidates as~\cite{he2018amc} in meta-learning.  We note that the speed gain over prior work~\cite{he2018amc} mainly comes from two sources: (i) generating the layer schedule with reinforcement learning requires $L$ network inferences while for our approach, we merely draw $L$ random variables from $L$ normal distributions and (ii) our evolutionary algorithm is non-parametric, which means we do not have to conduct backpropagation for learning while it must be performed through the controller network in the reinforcement learning approach. Additionally, with our formulation in equation (\ref{eq:latent}), we do not have to constrain the solution space, \emph{i.e.}, $\beta$, while the formulation in prior art~\cite{he2018amc} requires the solution space for the later layers to be constrained to satisfy the overall resource constraints.

\begin{figure}[h]
    \centering
    \includegraphics[width=\linewidth]{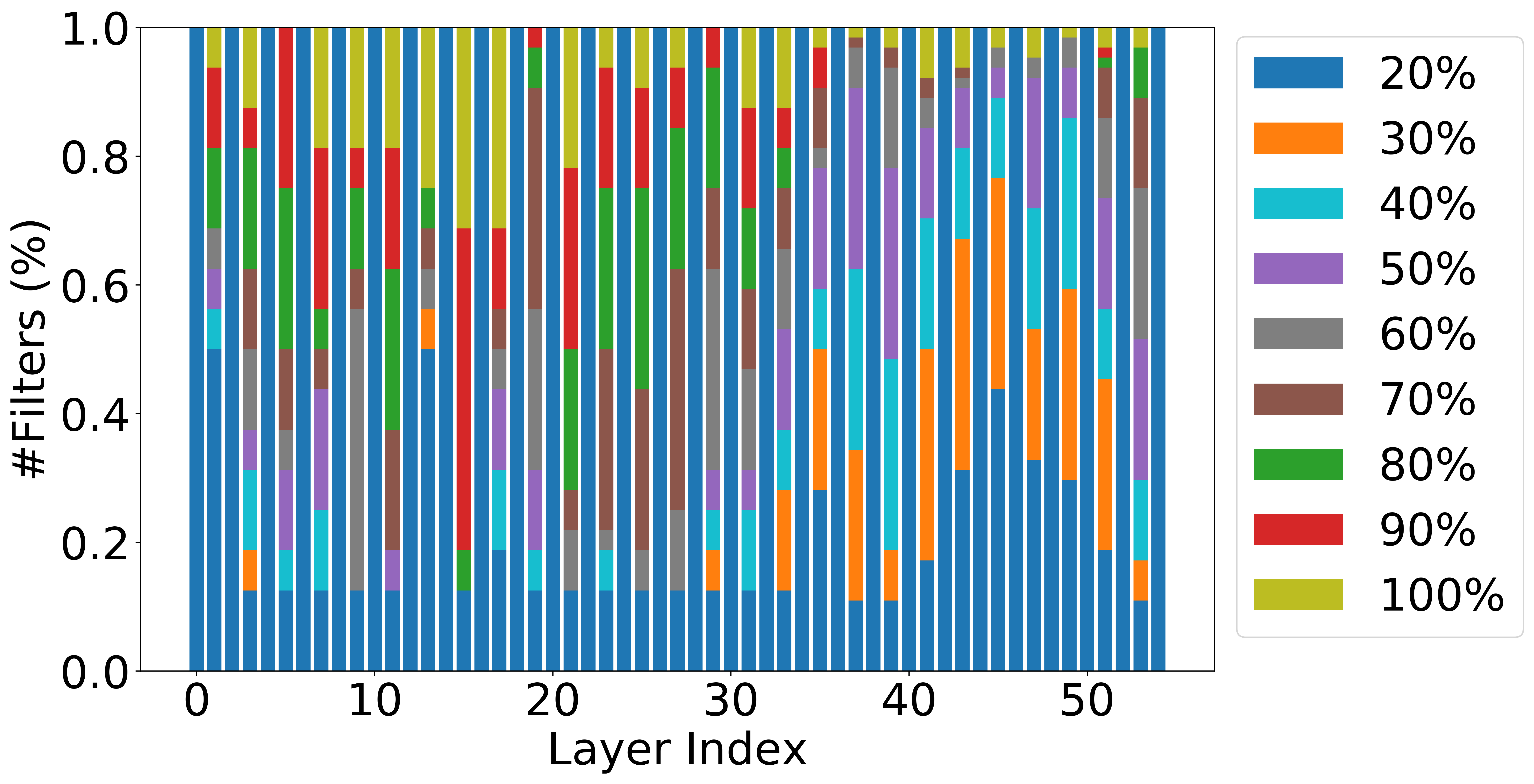}
    \caption{The layer schedule produced by the na\"ive pruning for different MAC constraint values. For example, the blue color bars represent the layer schedule when the network is pruned to 20\% MAC operations.}
    \label{fig:schedule}
\end{figure}

\paragraph{Na\"ive Pruning}
One might wonder why the na\"ive pruning works well under the MAC operations constraint, and hence, we visualize the layer schedules produced by the na\"ive pruning. As shown in Fig.~\ref{fig:schedule}, we plot the number of filters for each layer, normalized to the original network, under different MAC constraint values. First, we note that the residual connections are not pruned although they are allowed to be; this is due to the fact that a lot of convolutions are grouped together by the residual connections and this results in large metric score (Section~\ref{sec:impl}). On the other hand, we observe that shallower layers get pruned earlier than deeper layers and since shallower layers consist of more MAC operations due to larger feature maps, pruning them earlier results in a better Pareto frontier under the MAC operations constraint. This provides some evidences toward the effectiveness of $\ell_2$-based na\"ive pruning under the MAC constraint and its worse performance under the model size constraint.

\begin{figure}[h]
    \centering
    \includegraphics[width=\linewidth]{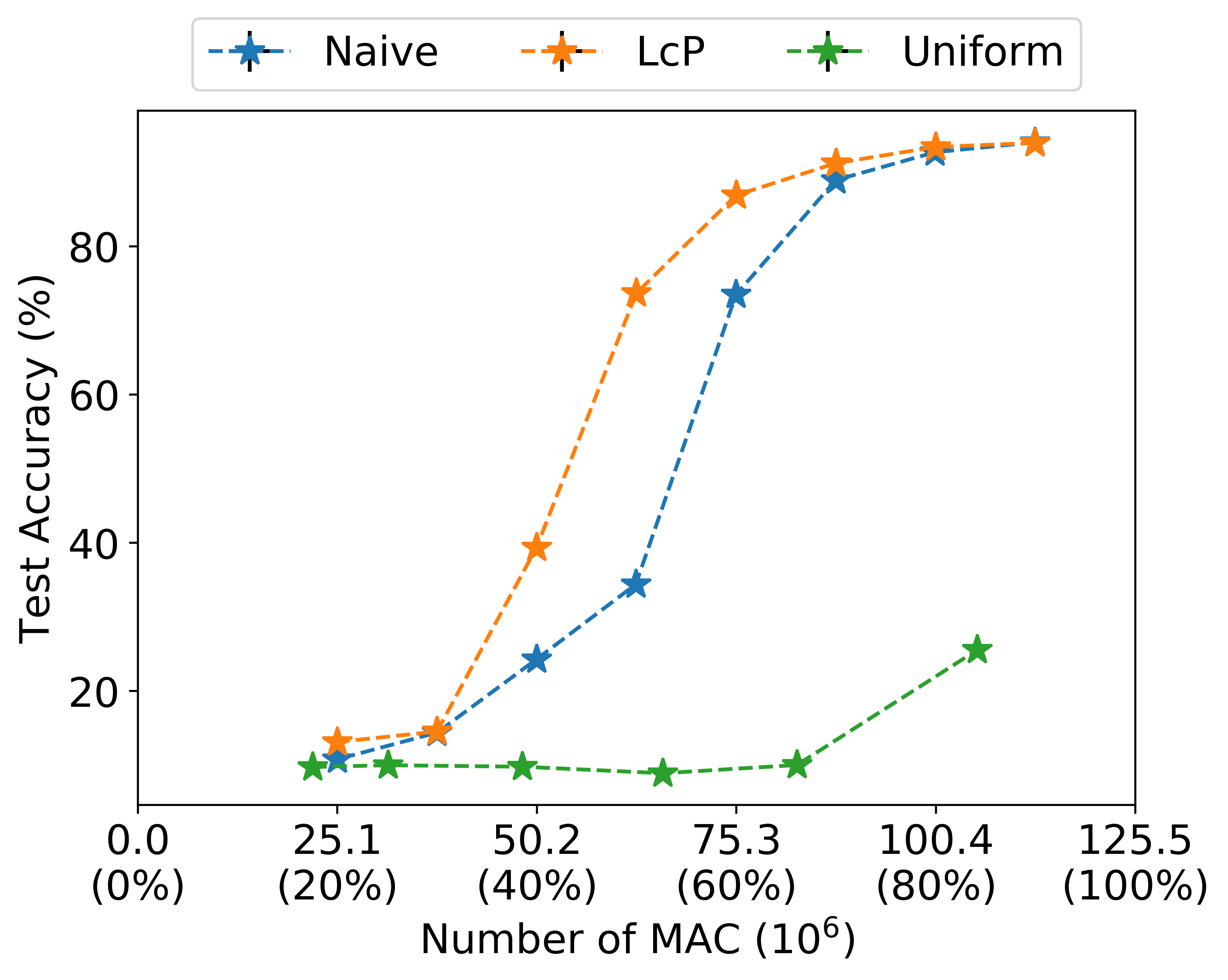}
    \caption{The testing accuracy of the pruned network before fine-tuning.}
    \label{fig:pretune}
\end{figure}

\paragraph{Layer-compensated Pruning}
To reason about the effectiveness of the proposed layer-compensated pruning (LcP), we plot the pre-tuning accuracy for the na\"ive, the uniform, and LcP. As we can see from Fig.~\ref{fig:pretune}, LcP produces networks with higher accuracy without fine-tuning, which means the latent variables that we find by solving equation (\ref{eq:latent}) using the training dataset do not over-fit to the training set. We conjecture that a good pre-tuning accuracy acts as a better initialization that leads to a better final performance after fine-tuning.

\begin{figure}[h]
    \centering
    \includegraphics[width=0.9\linewidth]{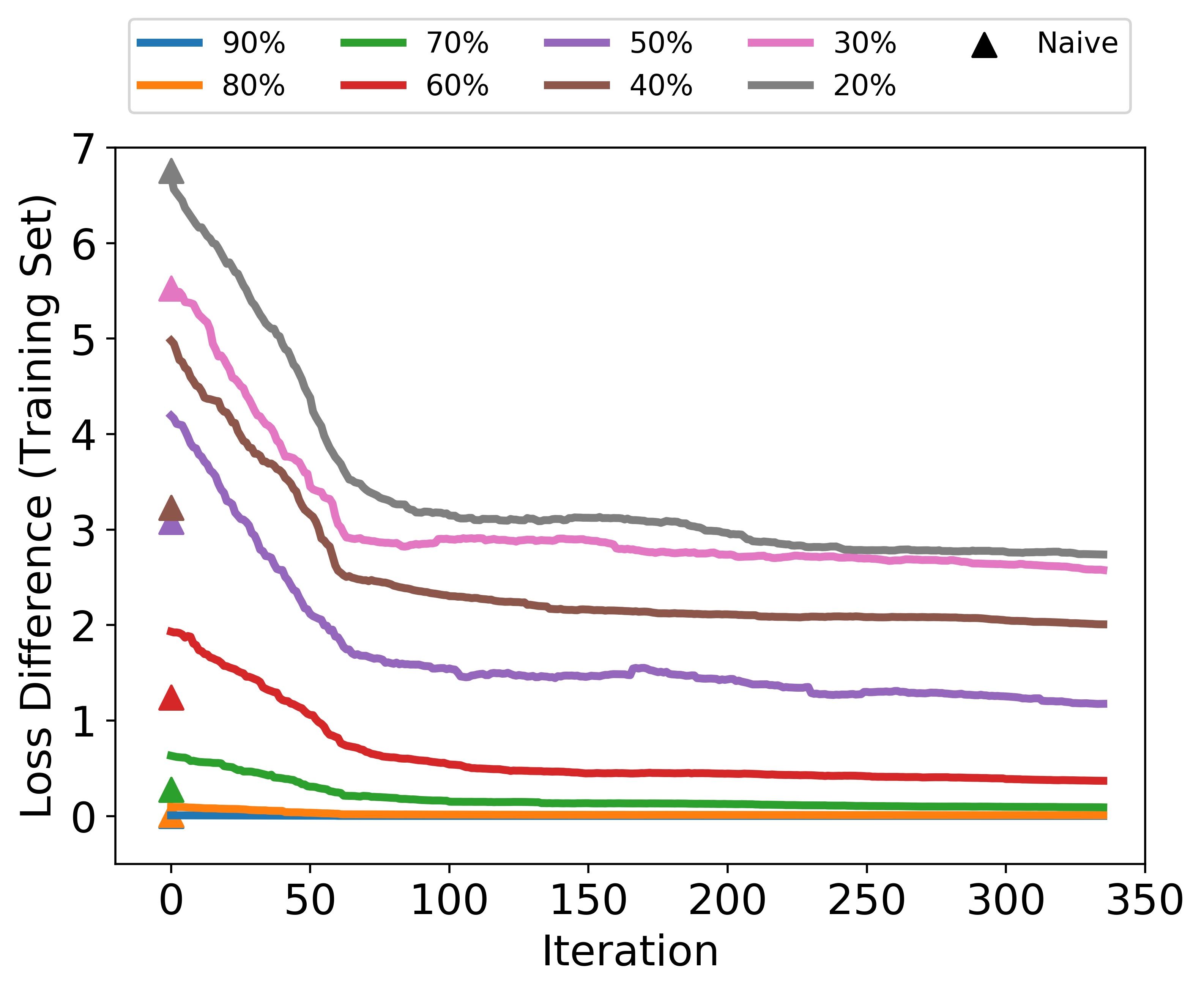}
    \caption{The training curves for the evolutionary algorithm solving equation (\ref{eq:latent}) with different values of the MAC operation constraint. For each constraint, we plot the loss difference for the na\"ive approach using a triangle in the corresponding color.}
    \label{fig:meta}
\end{figure}

We further plot the training curve of our evolutionary algorithm when solving equation (\ref{eq:latent}) targeting different MAC operations constraints in Fig.~\ref{fig:meta} to understand the process of the meta-learning. We find that when the targeting constraint values are loose, \emph{i.e.}, 90\% to 80\% of the original MAC operations, the improvement brought by the latent variables is not significant since it is easier to prune the network by just a little without accuracy degradation. On the other hand, our algorithm is more preferable compared to the na\"ive solution as the pruning constraint gets more stringent. Additionally, the algorithm converges around 100 iterations, which implies that either the step size $\alpha$, as mentioned in Section~\ref{sec:lc}, could be further optimized or one could perform an early stopping.

\begin{figure}[h]
    \centering
    \includegraphics[width=\linewidth]{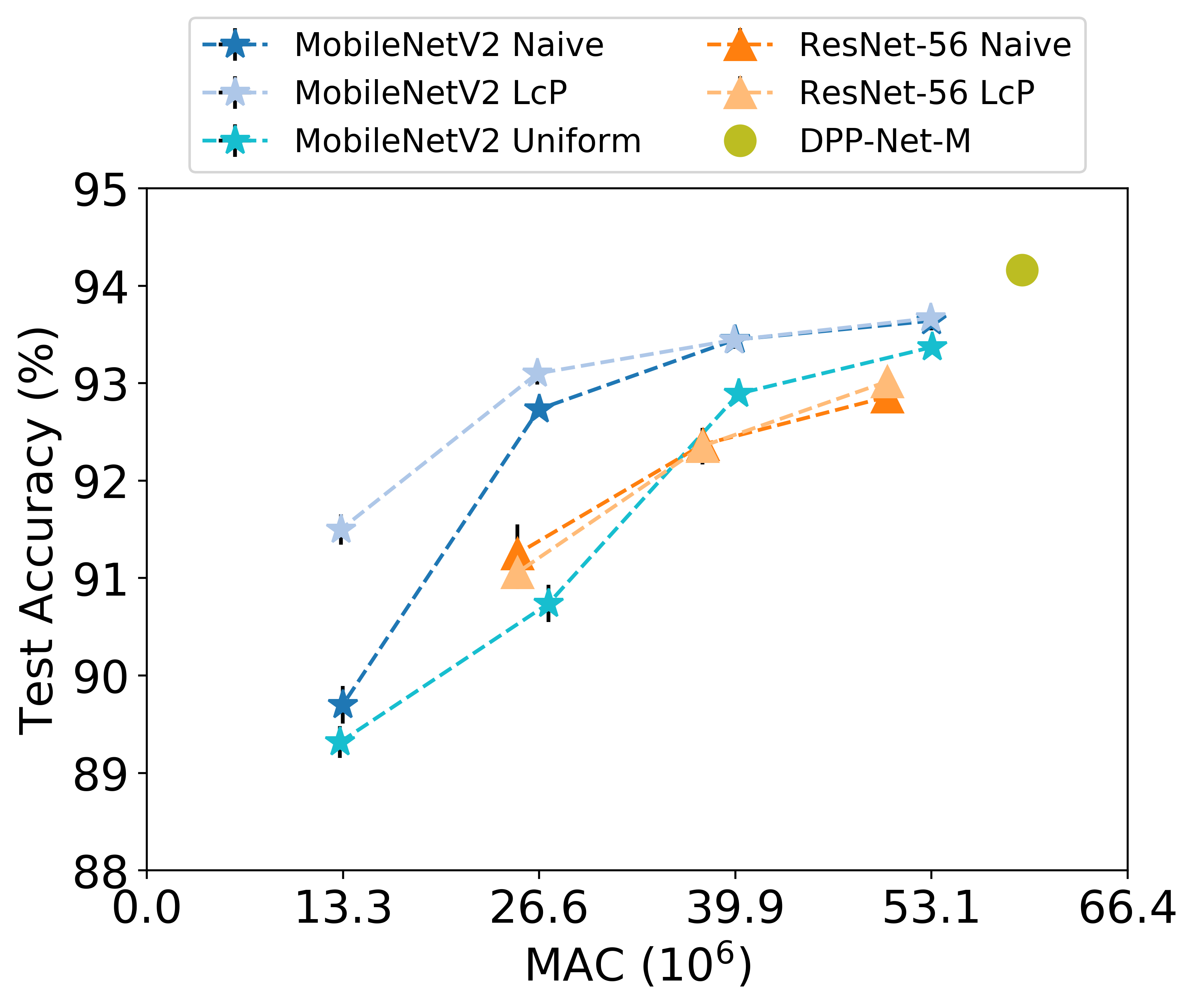}
    \caption{The networks characterized by small number of MAC operations. We plot the mean and standard deviation over three trials.}
    \label{fig:pareto_smallnet}
\end{figure}

\begin{figure}[h]
    \centering
    \includegraphics[width=\linewidth]{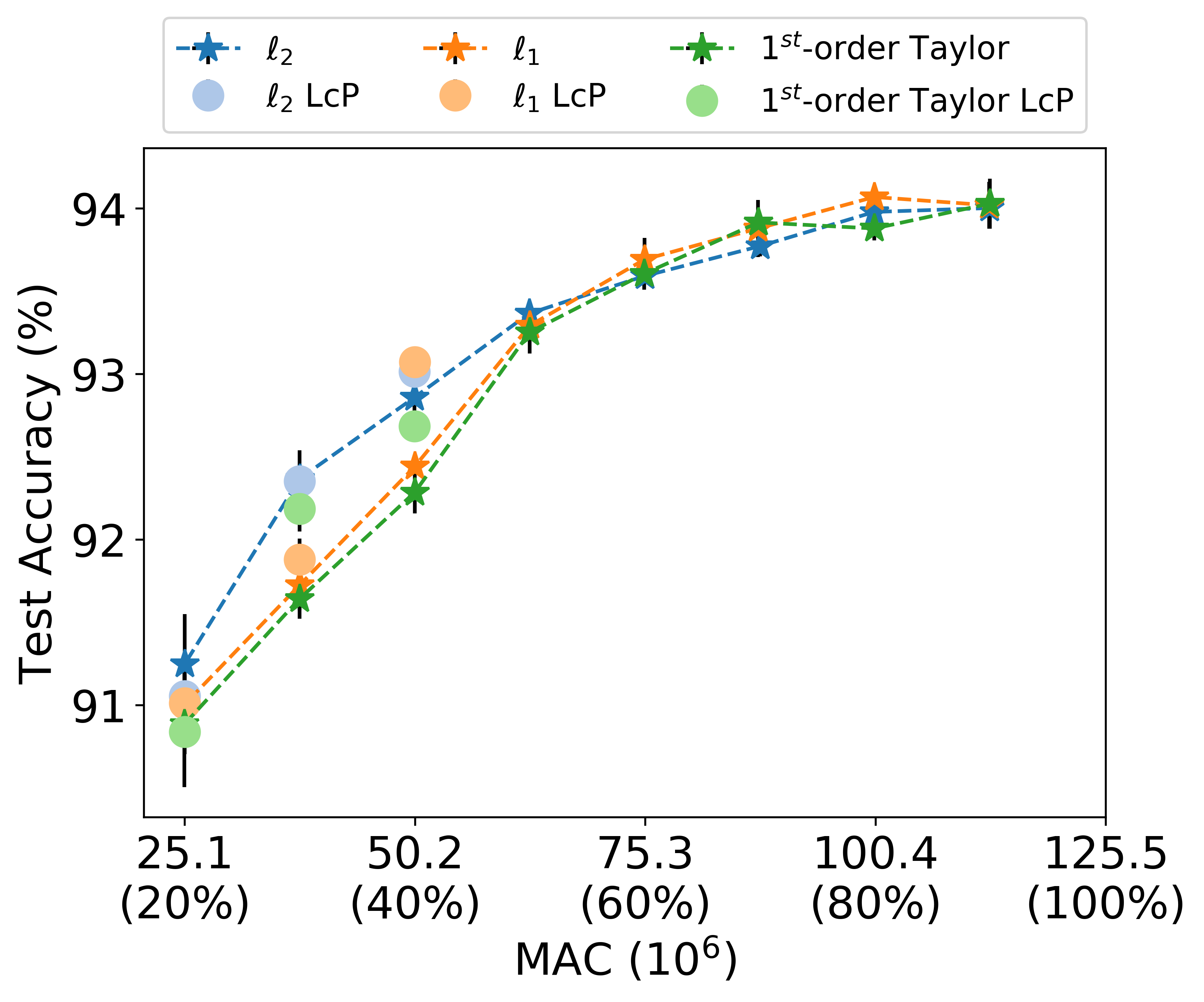}
    \caption{Pruning ResNet-56 with na\"ive pruning on CIFAR-10 using various of heuristic metrics.}
    \label{fig:metric}
\end{figure}

\paragraph{Compact Networks}
Other than ResNet-56, we also try MobileNetV2 on CIFAR-10. As shown in Fig.~\ref{fig:pareto_smallnet}, we plot the networks with small MAC operations including a network that is obtained by neural architectural search~\cite{dong2018dpp}. We first note that the 5x theoretical speedup (53.1M MAC operations) is achieved without accuracy degradation for MobileNetV2. Aligning with ResNet-56, the simple na\"ive pruning determines a better Pareto frontier compared to the uniform pruning while LcP performs the best. In particular, LcP achieves 2\% accuracy improvement (statistically significant) compared to both na\"ive and uniform at 13.3M MAC operations. Additionally, we find that with LcP, we can push existing networks closer to the network obtained by neural architecture search (\emph{i.e.}, DPP-Net-M). Also, it is worth noting that ResNet-56 is able to compete with uniformly pruned MobileNetV2 when pruned with both the na\"ive and LcP.
\paragraph{Different Heuristic Metrics}
Different heuristic metrics result in different approximation errors and hence, different performance in the pruned network. As shown in Fig.~\ref{fig:metric}, we find that with LcP, the performance of all considered heuristic metrics could be further improved.

\subsection{ImageNet Results}
For ImageNet, we conduct iterative constraint tightening. We gradually prune away 25\%, 42\%, and 50\% MAC operations. We note that iterative pruning is adopted in prior work for ImageNet as well~\cite{he2018amc}. We demonstrate the effectiveness of LcP on ResNet-50. As shown in Table~\ref{table:imagenet}, LcP is superior to prior art that reports on ResNet-50, which aligns with our observation in previous analysis. When pruning away 25\% of MAC operations, LcP achieves even higher accuracy compared to the original one. Furthermore, when pruned to 58\% and 50\% our algorithm achieves state-of-the-art results compared to prior art. Since the number of MAC operations does not directly translate into speedup, we report the latency of the pruned network with various inference batch size as shown in Table~\ref{table:latency}.
\begin{table}[h]
\caption{Summary of pruned ResNet-50 on ImageNet.}
\vskip 0.15in
\begin{center}
\begin{small}
\begin{sc}
\begin{adjustbox}{max width=0.46\textwidth}
\begin{tabular}{cccccc}
\toprule
 Method & Top-1 & Top-1 Diff & Top-5 & MAC (\%)\\
\midrule
LcP (Ours) & 76.13 $\xrightarrow{}$ 76.22 & +0.09 & 92.86 $\xrightarrow{}$ \textbf{93.05} & 75\\
\cite{Yu_2018_CVPR} & - $\xrightarrow{}$ - & -0.21 & - $\xrightarrow{}$ - & 73\\
\midrule
\cite{Huang_2018_ECCV} & 76.12 $\xrightarrow{}$ 74.18 & -1.94 & 92.86 $\xrightarrow{}$ 91.91 & 69\\
\cite{luo2017thinet} & 72.88 $\xrightarrow{}$ 72.04 & -0.84 & 91.14 $\xrightarrow{}$ 90.67 & 63\\
\cite{lin2018accelerating} & 75.13 $\xrightarrow{}$ 72.61 & -2.52 & 92.30 $\xrightarrow{}$ 91.05 & 58\\
\cite{he2018soft} & \textbf{76.15} $\xrightarrow{}$ 74.61 & -1.54 & 92.87
$\xrightarrow{}$ 92.06 & 58\\
LcP (Ours) & 76.13 $\xrightarrow{}$ \textbf{75.28} & -0.85 & 92.86 $\xrightarrow{}$ \textbf{92.60} & 58\\
\cite{Yu_2018_CVPR} & - $\xrightarrow{}$ - & -0.89 & - $\xrightarrow{}$ - & 56\\
\midrule
\cite{he2017channel} & - $\xrightarrow{}$ - & - & 92.2 $\xrightarrow{}$ 90.8 & 50\\
\cite{wang2017structured} & - $\xrightarrow{}$ - & - & 91.2 $\xrightarrow{}$ 90.4 & 50\\
LcP (Ours) & 76.13 $\xrightarrow{}$ \textbf{75.17} & -0.96 & 92.86 $\xrightarrow{}$ \textbf{92.44} & 50\\
\bottomrule
\end{tabular}\label{table:imagenet}
\end{adjustbox}
\end{sc}
\end{small}
\end{center}
\vskip -0.1in
\end{table}

\begin{table}[h]
\caption{Latency profile of ResNet-50. We report the latency per image in millisecond for different batch sizes (BS).}
\vskip 0.15in
\begin{center}
\begin{small}
\begin{sc}
\begin{adjustbox}{max width=0.46\textwidth}
\begin{tabular}{ccccc}
\toprule
 Method & $BS=1$ & $BS=4$ & $BS=16$ & $BS=64$\\
\midrule
ResNet50 & 5.03 & 2.17 & 1.53 & Out of Memory\\
ResNet50 (2x) & 4.62 & 1.69 & 1.11 & 0.99\\
\bottomrule
\end{tabular}\label{table:latency}
\end{adjustbox}
\end{sc}
\end{small}
\end{center}
\vskip -0.1in
\end{table}

\subsection{Transfer Learning Results}
We analyze how the LcP performs in a transfer learning setting where we have a model pre-trained on a large dataset, \emph{e.g.}, ImageNet, and we want to transfer its knowledge to adapt to a smaller dataset, \emph{e.g.}, Bird-200. We prune the fine-tuned network on the target dataset directly instead of pruning on the large dataset before transferring for two reasons: (i) the user only cares about the performance of the network on the target dataset instead of the source dataset, which means we need the Pareto frontier in the target dataset and (ii) pruning on a smaller dataset is much more efficient compared to pruning on a large dataset. We first obtain a fine-tuned MobileNetV2 on the Bird-200 dataset with top-1 accuracy 80.22\%, which matches the reported number from VGG-16 as well as DenseNet-121 from prior art~\cite{Mallya_2018_CVPR}. With an already small model such as MobileNetV2, we are able to achieve 78.34\% accuracy with 51\% MAC operations (153M) while our implemented greedy single-filter pruning~\cite{molchanov2016pruning} achieves 75.94\% with 50\% MAC operations~(150M).
\section{Ablation Study}
\subsection{Limiting the Layer Schedule}\label{sec:limit}
\begin{figure}[t]
    \centering
    \includegraphics[width=\linewidth]{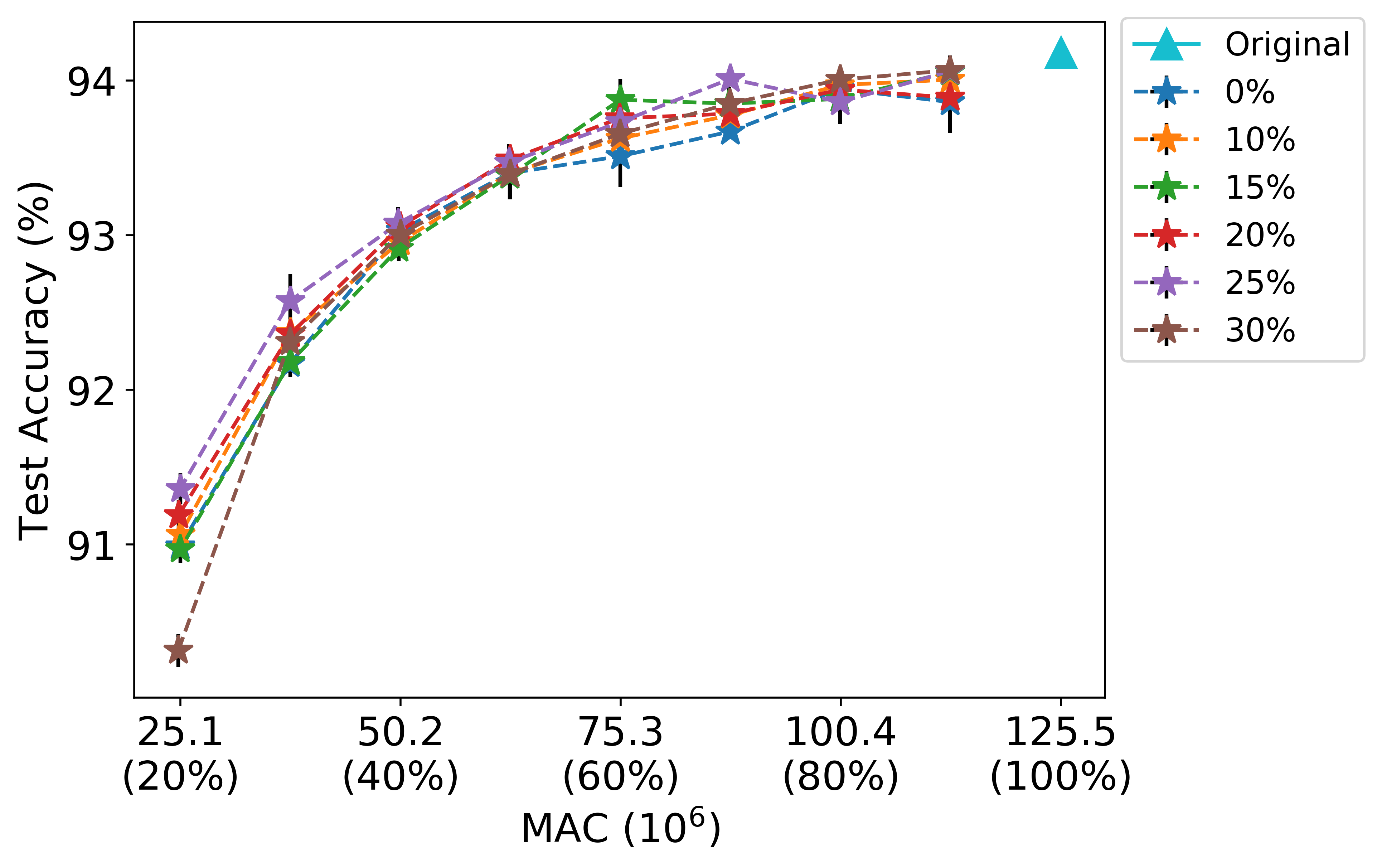}
    \caption{Pruning ResNet-56 with na\"ive pruning on CIFAR-10 using various of pruning budgets.}
    \label{fig:budget}
\end{figure}
Since we arbitrarily pick 10\% just to intuitively avoid extreme pruning, we study the effect of different pruning budgets on the performance of na\"ive pruning. As shown in Fig.~\ref{fig:budget}, we find that different budgets have similar performance. However, we observe a drop for 30\% budget on a 20\% MAC constraint. It is because pruning down to 20\% MAC operations results in similar a layer schedule produced by the uniform pruning when the budget is high, \emph{i.e.,} 30\%. We note that although CIFAR-10 performs fine even without budgeting, we find that such budgeting is needed when pruning for the Bird-200 dataset.
\section{Conclusion}
In this work, we consider the filter pruning problem as a global ranking problem rather than two separate sub-problems as commonly used in the literature. Moreover, based on the analysis of the approximation error incurred from the simplification of the problem, we propose the \emph{layer-compensated pruning} (LcP) that uses meta-learning to learn a set of latent variables that compensate for the layer-wise approximation error and it is able to improve the performance for various heuristic metrics. With such a formulation, we can learn the layer schedule with slightly better performance using \textbf{8x} less time compared to the reinforcement learning approach proposed in prior art, which is significant if one considers Pareto frontier traversal. Moreover, targeting networks with small number of MAC operations, our algorithm produces networks comparable with the network determined by a bottom-up approach while being superior to the uniform and na\"ive pruning. Last, we conduct comprehensive analysis on the proposed method to demonstrate both the effectiveness and the efficiency of our approach using two types of neural networks and three datasets.


\bibliography{example_paper}
\bibliographystyle{sysml2019}

\end{document}